# Investigations on convergence behaviour of Physics Informed Neural Networks across spectral ranges and derivative orders

Mayank Deshpande, Siddharth Agarwal, Vukka Snigdha, Arya Kumar Bhattacharya*

*Department of Computer Science, Mahindra University, Hyderabad, India*

*Arya.Bhattacharya@mahindrauniversity.edu.in* *

*Abstract*—An important inference from Neural Tangent Kernel (NTK) theory is the existence of spectral bias (SB), that is, low frequency components of the target function of a fully connected Artificial Neural Network (ANN) being learnt significantly faster than the higher frequencies during training. This is established for Mean Square Error (MSE) loss functions with very low learning rate parameters. Physics Informed Neural Networks (PINNs) are designed to learn the solutions of differential equations (DE) of arbitrary orders; in PINNs the loss functions are obtained as the residues of the conservative form of the DEs and represent the degree of dissatisfaction of the equations. So there has been an open question whether (a) PINNs also exhibit SB and (b) if so, how does this bias vary across the orders of the DEs. In this work, a series of numerical experiments are conducted on simple sinusoidal functions of varying frequencies, compositions and equation orders to investigate these issues. It is firmly established that under normalized conditions, PINNs do exhibit strong spectral bias, and this increases with the order of the differential equation.

## I. INTRODUCTION

Coupled sets of Partial Differential Equations (PDEs), like the Navier-Stokes equations for Fluid Mechanics [1], Maxwell's equations for Electromagnetics [2], and analogous equations in other domains, govern the behavior and dynamic field characteristics in their respective areas. In most practical applications these domains turn out to be quite complicated and not amenable to closed form solutions. Numerical simulations of these equations, using either the Integral Equation approach [3][4] that are based fundamentally on the Green's Identities [5], or the more popular discretized form approaches [6] [7], under appropriate boundary and initial conditions, have performed till date as the best-known techniques for solution of these equations to high levels of fidelity suitable to the relevant application.

Recently, the Universal Approximation capabilities of Artificial Neural Networks [8][9] have been extended to the solution of sets of PDEs under the framework of Physics Informed Neural Networks [10] and in a short period this approach has seen rapid growth both in theory and applications in multiple domains. Particularly in the context of quick turnaround times for solutions with minor changes in design and flow conditions, Physics Informed Neural Networks (PINNs) have already demonstrated their utility, see e.g. [11].

PINNs represent a specialized approach within the gamut of Artificial Neural Network (ANN) architectures and techniques, and it may be expected that the properties of ANNs will apply onto PINNs as well. Using the theory of Neural Tangent Kernel [12] (NTK) it can be mathematically deduced [13] that in the training process of a fully-connected multi-layer ANN with Mean Square Error (MSE) as the loss function, and with wide hidden layers and a very small learning rate parameter, the lower frequency components of the function being learnt will converge faster, and the higher frequencies will be acquired more slowly. That is, higher the frequency component, the slower it will be learnt. This phenomenon is known as Spectral Bias [14].

Now, PINNs do not use MSE and are not based on supervised learning, but instead the *residues* of the equations in the field and at the boundaries perform the role of the loss function. So, it has been an open question if the deductions from NTK theory that map into the phenomena of spectral bias in ANNs trained with standard supervised learning approach, will be valid for PINNs.

This ambiguity on the existence of spectral bias in PINNs has led to minor speculation along the following lines:

(a) if solution of functions represented as differential equations will indeed exhibit spectral bias in the learning process [15]

(b) and, if (a) is true, if the extent of variation in the rate of convergence across frequency components will increase or reduce with the order of the differential equations.

This work seeks to answer the above questions conclusively through a series of numerical experiments. In addition, it also investigates:

(c) the extent to which spectral bias varies with different activation functions, and

(d) for functions without derivatives (i.e., not expressed as differential equations) that can be solved by conventional ANNs using supervised learning and training data, the extent of difference in overall convergence rates and spectral bias between solutions obtained by the conventional approach, versus solutions obtained using a PINN framework.

The rest of this paper is organized as follows. Section II provides a brief theoretical background of PINNs. Section III provides the summary of the concept of spectral bias and its deduction from NTK theory. Section IV provides the results of numerical experimentation, where different sub-sections discuss the functions being tested and the corresponding results and observations. Conclusions are discussed in Section V.

## II. BASIC FORMULATION OF PHYSICS INFORMED NEURAL NETWORKS

We begin with a statement of the general form of sets of PDEs, as typically applicable to Fluid Mechanics and other domains with analogous sets of governing partial differential equations

$$N_i[u](x) = f_i(x), \quad \forall i \in \{1,...,N_N\}, \quad x \in D, \tag{1}$$

$$C_j[u](x) = g_j(x), \quad \forall j \in \{1,...,N_C\}, \quad x \in \partial D, \tag{2}$$

where $N_i[.]$ are general differential operators that are applied on functions $u(x)$, where $x$ is the set of independent location vectors defined over a bounded continuous domain $D \subseteq \Box^d$, $d \in \{1,2,3,....\}$, $i$ represents the number of operators corresponding to equations in the set, $u$ is a vector of dependent variables of interest representing the solution at the field points $x$. $C_j[.]$ denote constraint operators that consist of differential, linear and non-linear terms and usually cover the boundary and initial conditions, $j$ denotes the set of constraint operators. $\partial D$ denotes a subset of the domain boundary that is needed to define the constraints. It may be noted that x can in principle denote both location and time variables, in which case (2) will extend to both boundary and initial conditions.

As an example, for 3D laminar flow described by the Navier-Stokes equations, the number of equations $N_N$ is 4, of which 3 are the momentum equations and one that of continuity. The boundary conditions represented by eq. (2) depend on the flow specifics; on solid boundaries these will be the zero normal and tangential flow conditions. In this framework the components of *u* at any *x* are the 3 velocity components and the pressure (and possibly density).

We seek to approximate the solution $u(x)$ by a neural network $u_{net}(x; \theta)$, where $\theta$ is the vector of parameters of the neural network, as typically represented in ML models. Indeed, if $u_{net}(x; \theta)$ were to precisely represent the solution, then (1) & (2) could be expressed as

$$N_i[u_{net}(\theta)](x) - f_i(x) = 0, \quad \forall i \in \{1, ..., N_N\}, \quad x \in D, \tag{3}$$

$$C_j[u_{net}(\theta)](x) - g_j(x) = 0, \quad \forall j \in \{1, ..., N_C\}, \quad x \in \partial D. \tag{4}$$

However, no ML model can generate precisely zero error, so it is reasonable to write the above equations in the form of residuals

$$r_N^{(i)}(x; u_{net}(\theta)) = N_i[u_{net}(\theta)](x) - f_i(x), \tag{5}$$

$$r_C^{(j)}(x; u_{net}(\theta)) = C_j[u_{net}(\theta)](x) - g_j(x), \tag{6}$$

in (5) and (6) the ranges of *i* and *j* and distribution of *x* are no longer restated for brevity. Also, that the residuals are expressed as functions of the current state of the network (i.e. its parameters).

The procedure for training of the ANN, i.e. attainment of the best possible parameters $\theta$ to enable $u_{net}(x; \theta)$ to represent (3) & (4) as closely as possible, is to minimize the net loss function

$$\mathcal{L}_{res}(\theta) = \sum_{i=1}^{N_N} \int_{\mathcal{D}} \lambda_N^{(i)}(\mathbf{x}) \|r_N^{(i)}(\mathbf{x}; u_{net}(\theta))\|_p d\mathbf{x} + \sum_{j=1}^{N_C} \int_{\partial \mathcal{D}} \lambda_C^{(j)}(\mathbf{x}) \|r_C^{(j)}(\mathbf{x}; u_{net}(\theta))\|_p d\mathbf{x} \tag{7}$$

where $\|\cdot\|_p$ denotes the p-norm, and $\lambda_N^{(i)}, \lambda_C^{(j)}$ are weight functions that control the loss interplay between the equation terms and the constraint terms, as well as across the different equation and constraint terms (represented as summations). It may be stated here that the values of these λ's play a crucial role in the accuracy and the convergence of the solutions and are subjects of ongoing research. The integration symbols do not really denote integration over continuous spaces, but Monte-Carlo integration over a fairly large number of points (cloud) selected in and on the respective volumes and surfaces.

The PINN mechanism is illustrated in the figure 1. below, avoiding detail.

### III. DEDUCTION OF SPECTRAL BIAS FROM NTK THEORY

This section first expresses the fundamental result from Neural Tangent Kernel Theory [12, 16] and from there deduces the varying convergence rates of different frequency components of the function being trained, for a conventional MSE-loss based ANN.

Let $f(x, \theta)$ represent a scalar valued fully connected ANN with weights $\theta$ initialized by a Gaussian distribution. Considering a training data set $\{X_{trn}, Y_{trn}\}$ composed of N samples, one may express inputs $X_{trn}$ as $(x_i)_{i=1}^N$ and the corresponding outputs $Y_{trn}$ as $(y_i)_{i=1}^N$. If the ANN is trained using the Mean Square Error loss

function

$$\mathbf{L}(\theta) = \frac{1}{N} \sum_{i=1}^{N} (f(x_i, \theta) - y_i)^2 \qquad (8)$$

with a very small value of learning rate parameter η, then, using the derivation of Jacot et al [12, 16], one may define the Neural Tangent Kernel (NTK) operator $\mathbf{K}$, with entries given by

$$\mathbf{K}_{i,j} = \mathbf{K}(x_i, x_j) = \left\langle \frac{\partial f(x_i, \theta)}{\partial \theta}, \frac{\partial f(x_j, \theta)}{\partial \theta} \right\rangle. \qquad (9)$$

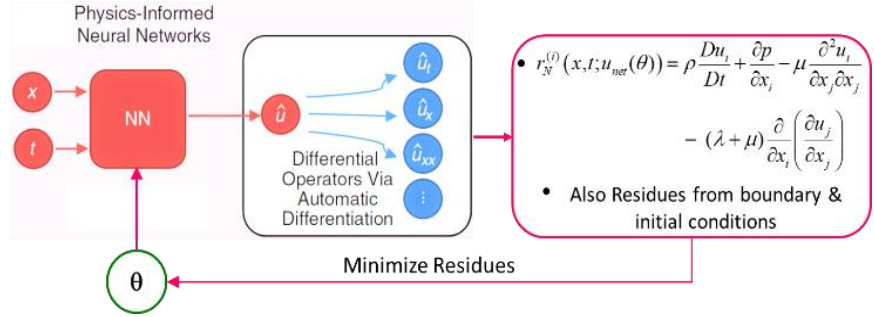

Fig. 1. PINN training process. The NN inputs are the points cloud $x$ at multiple time steps. The outputs are the values of vector $u_{net}$ or $\hat{u}$ at each of the input points. Then *Automatic Differentiation* is used to evaluate each of the derivative terms in all the PDEs and Constraints, which are finally converted into the different components of the Loss as in eq. (7). Residue shown of NS equations.

The NTK theory shows that, under the above conditions and using a gradient descent approach to training, the kernel $\mathbf{K}$ converges to a deterministic value and does not change even if the width of the network hidden layer/s increase towards infinity.

Further, it can be shown that [17]

$$\frac{df(\mathbf{X}_{trn}, \theta(t))}{dt} \approx -\mathbf{K} \cdot \left( f(\mathbf{X}_{trn}, \theta(t)) - \mathbf{Y}_{trn} \right) \qquad (10)$$

where $\theta(t)$ denotes the parameters of the network at iteration $t$; the vector form of differential equation (10) may be observed. Solution of (10) may be expressed as

$$f(\mathbf{X}_{trn}, \theta(t)) \approx (\mathbf{I} - e^{-\mathbf{K}t}) \mathbf{Y}_{trn} \qquad (11)$$

The kernel $\mathbf{K}$ being square symmetric and positive semi-definite, we can express its spectral decomposition as

$$\mathbf{K} = \mathbf{Q}\mathbf{\Lambda}\mathbf{Q}^T \qquad (12)$$

where $\mathbf{Q}$ is an orthogonal matrix with $i^{th}$ column as the eigenvector $q_i$ of $\mathbf{K}$, and $\mathbf{\Lambda}$ is a diagonal matrix with entries $\lambda_i$ as the corresponding eigenvalues. Also note that $\mathbf{Q}^T = \mathbf{Q}^{-1}$, and as

$$e^{-\mathbf{K}t} = \mathbf{Q} e^{-\mathbf{\Lambda}t} \mathbf{Q}^T \qquad (13)$$

From (11), one may write

$$\left( f(\mathbf{X}_{trn}, \theta(t)) - \mathbf{Y}_{trn} \right) \approx -e^{-\mathbf{K}t} \mathbf{Y}_{trn} \qquad (14)$$

where and on substituting from (13), (14) yields

$$(f(X_{trn}, \theta(t)) - Y_{trn}) \approx -Q e^{-\Lambda t} Q^T Y_{trn}$$

which can be further written as

$$Q^T (f(X_{trn}, \theta(t)) - Y_{trn}) \approx -e^{-\Lambda t} Q^T Y_{trn} \tag{15}$$

Equation (15) can be written in expanded form as shown below. Eq. (16) shows that the i$^{th}$ component of the absolute error, $|q_i^T \cdot (f(X_{trn}, \theta(t)) - Y_{trn})|$, will decay approximately exponentially at the rate $\lambda_i$. That is, components of the target function that correspond to kernel eigenvectors with larger eigenvalues, will be learnt faster. The larger eigenvalues correspond to the larger spectral wavelengths and hence the smaller (lower) frequencies, and vice-versa, see e.g. [18] and [19]. Thus, for a fully connected ANN with MSE loss function and small learning rate parameter, in the process of training *the lower frequency components of the target function learn faster than the higher ones*.

$$\begin{bmatrix} q_1^T \\ q_2^T \\ q_3^T \\ . \\ . \\ q_N^T \end{bmatrix} (f(X_{trn}, \theta(t)) - Y_{trn}) = \begin{bmatrix} e^{-\lambda_1 t} & & & & \\ & e^{-\lambda_2 t} & & & \\ & & e^{-\lambda_3 t} & & \\ & & & . & \\ & & & & . \\ & & & & & e^{-\lambda_N t} \end{bmatrix} \begin{bmatrix} q_1^T \\ q_2^T \\ q_3^T \\ . \\ . \\ q_N^T \end{bmatrix} Y_{trn} \tag{16}$$

The question arises – does the above also hold for PINNs, where the loss function is different, and if yes, then how does this spectral bias vary across the orders of the differential equation that represents the target function? We perform a series of numerical experiments to address the above and related questions, and the results are reproduced and discussed in Section IV.

## IV. NUMERICAL EXPERIMENTATION AND RESULTS

*A. Equations of combined frequency terms*

A series of numerical experiments are performed on different sinusoidal functions represented as differential equations of various orders. The considered equations are represented below:

$$f(x) = \sum_{k=1}^{5} \sin(2kx)/(2k), \quad x \in [-\pi, \pi] \tag{17}$$

$$\frac{df(x)}{dx} = \sum_{k=1}^{5} \cos(2kx), \quad x \in [-\pi, \pi] \tag{18}$$

$$\frac{d^2 f(x)}{dx^2} = -\sum_{k=1}^{5} (2k) \sin(2kx), \quad x \in [-\pi, \pi] \tag{19}$$

$$\frac{d^3 f(x)}{dx^3} = -\sum_{k=1}^{5} (2k)^2 \cos(2kx), \quad x \in [-\pi, \pi] \tag{20}$$

It can be seen that eqns. (18-20) are increasing order differential equations of the function represented by (17), when the boundary conditions are set at $f(x) = 0$ at both the ends of the given range.

PINN solutions are obtained on the Nvidia Modulus framework [20] on the DGX-1 computing platform. Computations are performed for each of the equations (17-20), and while the typical plots of convergence and accuracy against the number of iterations are made, it is considered more pertinent here to show the plots of the developing solution (red curve) against the closed form solution, in blue. The plots are made for every

1000 (referred as 1 K) iterations till convergence, and only the plots at 1 K and 5 K iterations are shown, in figs. 2-7 for the eqns. (17-19). The third derivative solution did not converge; hence plots are not shown. When convergence occurs, the red and blue curves will coincide. The variation in convergence across the different derivative orders may be observed.

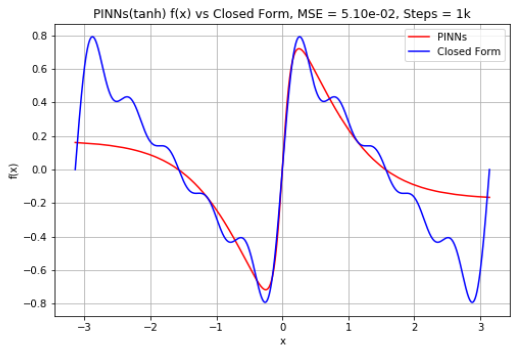

Fig. 2. Developing solution from PINN for $f(x)$, eq. 17, at 1 K iters (red) against the closed form solution (blue).

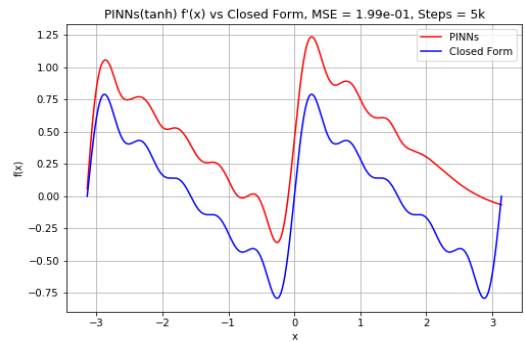

Fig. 5. Developing solution from PINN for $f_x(x)$, eq. 18, at 5 K iters (red) against the closed form solution (blue). One may compare against fig. 3 for $f(x)$. It appears that $f_x(x)$ is developing faster.

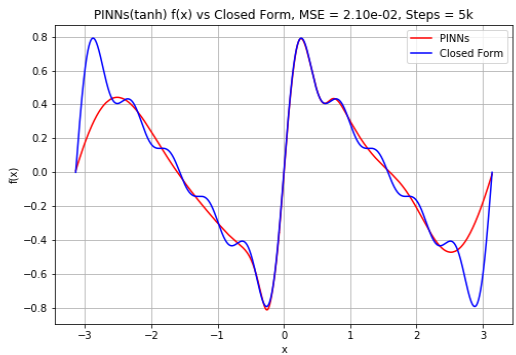

Fig. 3. Developing solution from PINN for $f(x)$, eq. 17, at 5 K iters (red) against the closed form solution (blue).

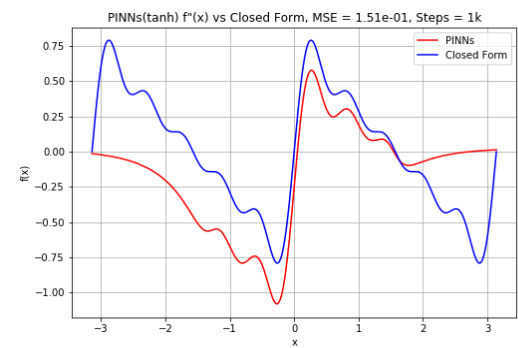

Fig. 6. Developing solution from PINN for $f_{xx}(x)$, eq. 19, at 1 K iters (red) against the closed form solution (blue). One may compare against fig. 2 & 4.

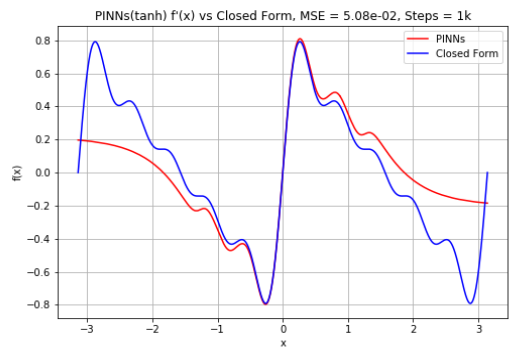

Fig. 4. Developing solution from PINN for $f_x(x)$, eq. 18, at 1 K iters (red) against the closed form solution (blue). One may compare against fig. 2 for $f(x)$.

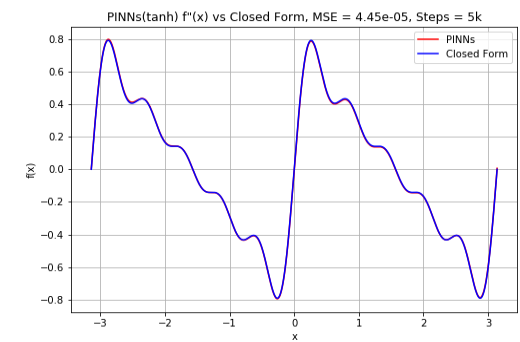

Fig. 7. Developing solution from PINN for $f_{xx}(x)$, eq. 19, at 5 K iters (red) against the closed form solution (blue). One may compare against fig. 3 for $f(x)$ and fig. 5 for $f_x(x)$. It is clear that $f_{xx}(x)$ has developed much faster. for $f(x)$ and fig. 5 for $f_x(x)$. It is clear that $f_{xx}(x)$ has developed much faster.

If one were to closely observe figs. 2-7, two important features can be seen. First, for any of the three equations for which solutions are obtained, the lower frequency components are captured relatively quickly, while the higher frequency components (the wiggles) are captured more slowly. Second, among the three equations, the highest order differential equation $f_{xx}(x)$ is resolved fastest, the intermediate derivative $f_x(x)$ slower, and the baseline function $f(x)$ slowest of all.

These aspects are captured in figs. 8-11 that show frequency-magnitude plots at different iteration levels, obtained by performing FFT on the solutions displayed above. Each figure shows the 5 relevant frequencies, and at each frequency, the magnitude of the 3 solutions against that of the closed form. The variant rates of evolution of the 3 solutions can be seen. each frequency, the magnitude of the 3 can be seen.

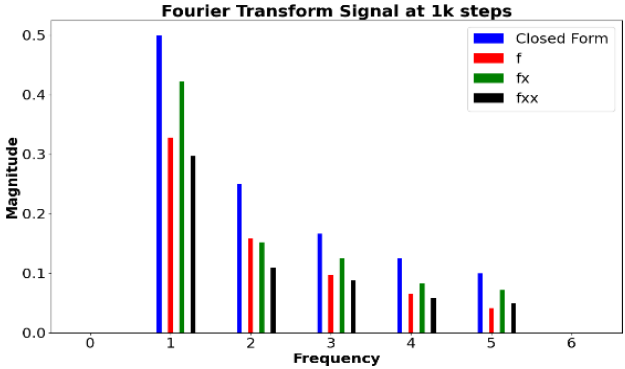

Fig. 8. Gradual development of different frequency components of $f(x)$, $f_x(x)$ and $f_{xx}(x)$ towards that of closed form frequency components. Here at 1 K, a clear picture is yet to emerge.

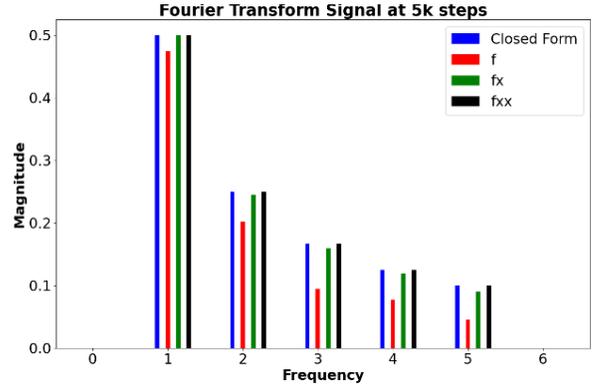

Fig. 9. Gradual development of different frequency components of $f(x)$, $f_x(x)$ and $f_{xx}(x)$ towards that of exact frequency components. Here at 5 K, $f_{xx}(x)$ frequency magnitudes have reached that of exact levels. $f_x(x)$ is close but $f(x)$ is still distant. Comparison can be made with fig. 7.

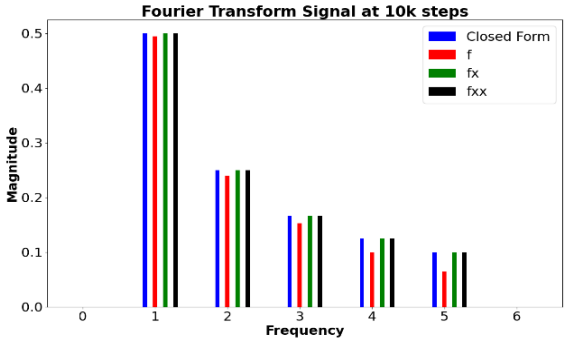

Fig. 10. Gradual development of different frequency components of $f(x)$, $f_x(x)$ and $f_{xx}(x)$ towards that of the exact solution. Here at 10 K, $f_x(x)$ frequency magnitudes have now reached that of exact levels. $f(x)$ is still at some distance. Noticeably, the higher frequencies are far from exact solutions; the lower frequencies are closer. The same trend is visible in fig.9.

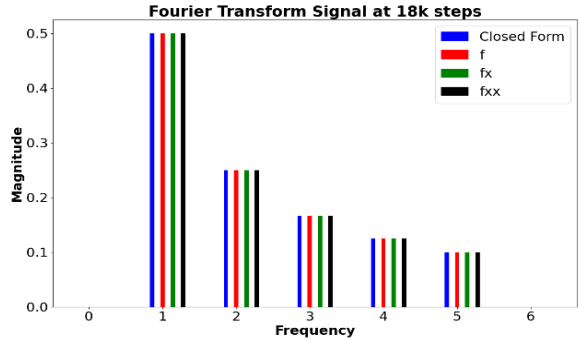

Fig. 11. At 18 K, the slowest evolving high frequency components of $f(x)$ have now reached that of exact levels. That is the point where $f(x)$ PINN solution has exactly coincided with the closed form, i.e. solution has converged.

From the above figures and observations, a clear pattern seems to emerge. Higher derivatives, i.e. differential equations of higher orders, seem to converge faster. *More specifically, the variation in rates of convergence across frequencies seems to reduce as we get to higher derivatives*. This *could possibly* have been a conclusion, except for the incompatible observation that when we get to the third order, the solution does not converge at all! Clearly, there is something else that is also affecting the process!

The above studies were conducted with tanh(.) as the activation function. A similar set of runs were made with the swish(.) activation function, here the results were analogous with a significant slowing down of the rates of convergence, compared to tanh(.), in all the cases. Further computations have been made with tanh(.) alone. Table 1 below summarizes all these results, without paying heed to the frequency aspects.

The combined frequency functions expressed as different order differential equations (17-20) are disadvantageous for frequency variation studies as the function plots cannot precisely discriminate between frequencies, and one has to take recourse to Fourier transforms to obtain the frequency-magnitude plots. Instead, now individual functions at different order derivatives are created with only single sinusoids of different frequencies, and analyzed in the next section.

Table 1. Convergences of functions of combined sinusoids

| Activation function | Number of iterations at which convergence is achieved for the function, in thousands (K) | | | |
| --- | --- | --- | --- | --- |
| | $f(x)$ (eq. 17) | $f_x(x)$, (eq. 18) | $f_{xx}(x)$, (eq. 19) | $f_{xxx}(x)$, (eq. 20) |
| tanh(.) | 18 | 10 | 5 | No convergence |
| swish(.) | 110 | 17 | 12 | No convergence |

*B. Functions with single frequency terms*

The functions considered are as below:

$$\frac{d^2 f(x)}{dx^2} = \sin kx, \text{ for k = 2, 6 and 10, } x \in [-\pi, \pi] \tag{21}$$

which is obtained from the following function:

$$f(x) = -\frac{1}{k^2} \sin kx, \text{ for k = 2, 6 and 10, } x \in [-\pi, \pi] \tag{22}$$

and $\frac{d^3 f(x)}{dx^3} = \cos kx$, for k = 2, 6 and 10, $x \in [-\pi, \pi]$ (23)

with its corresponding baseline function

$$f(x) = -\frac{1}{k^3} \sin kx, \text{ for k = 2, 6 and 10, } x \in [-\pi, \pi] \tag{24}$$

All of these functions are solved using the PINN approach within the Modulus framework. First, the results in terms of number of iterations for convergence for each of the frequency cases are presented in the table below:

Table 2. Convergences of functions of single sinusoids at different frequencies, activation function tanh(.)

| Frequency, k | Number of iterations at which convergence is achieved for the function, in thousands (K) | | |
| --- | --- | --- | --- |
| | $f(x)$ (eq. 22) | $f_{xx}(x)$, (eq. 21) | $f_{xxx}(x)$, (eq. 23) |
| 2 | 6 | 2 | 11 |
| 6 | 246 | 18 | 81 |
| 10 | No convergence | 50 | No convergence |

Table 2 clearly shows that low frequency cases are converging faster, irrespective of derivative order. However, there is no clear pattern across derivative orders for a given frequency, and the second order differential equation converges faster than either the third order or the baseline function. To obtain a better insight into the rates of convergence across frequencies and derivative orders, figs. 12-20 plot the function values for all 9 cases, 3 derivative orders at 3 different frequencies. Again, in each case the function at 5 K iterations (red) is plotted against the closed form solution (blue).

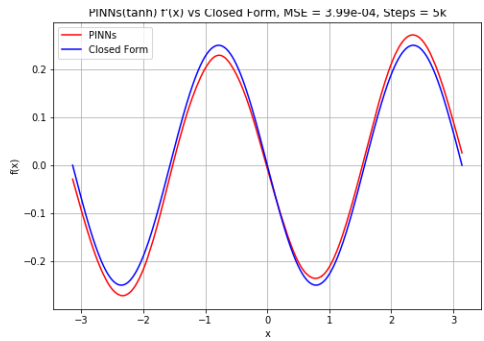

Fig. 12. PINN solution at 5 K iterations of $f(x)$ (eq. (22)) (red) against exact solution (blue), when frequency $k$ is 2.

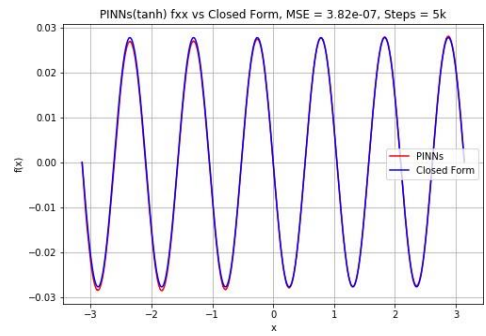

Fig. 16. PINN solution at 5 K iterations of $f_{xx}(x)$ (eq. (21)) (red) against exact solution (blue), when frequency $k$ is 6.

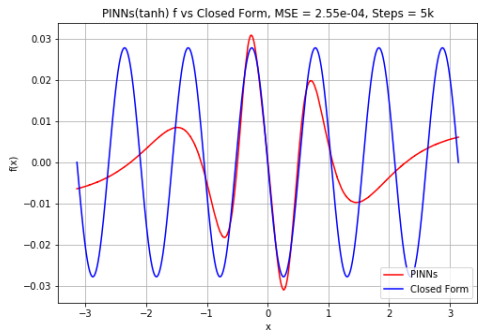

Fig. 13. PINN solution at 5 K iterations of $f(x)$ (eq. (22)) (red) against exact solution (blue), when frequency $k$ is 6.

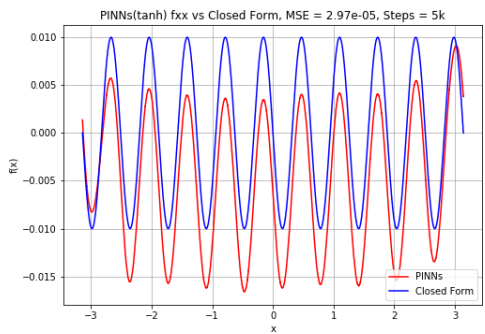

Fig. 17. PINN solution at 5 K iterations of $f_{xx}(x)$ (eq. (21)) (red) against exact solution (blue), when frequency $k$ is 10.

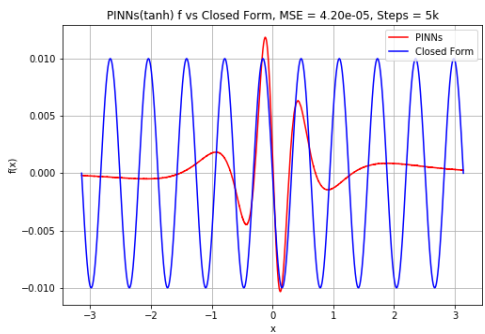

Fig. 14. PINN solution at 5 K iterations of $f(x)$ (eq. (22)) (red) against exact solution (blue), when frequency $k$ is 10. Note this case did not converge at all.

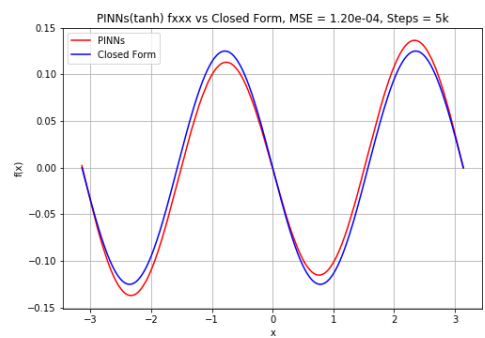

Fig. 18. PINN solution at 5 K iterations of $f_{xxx}(x)$ (eq. (23)) (red) against exact solution (blue), when frequency $k$ is 2.

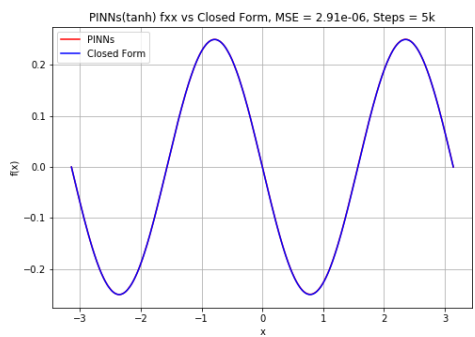

Fig. 15. PINN solution at 5 K iterations of $f_{xx}(x)$ (eq. (21)) (red) against exact solution (blue), when frequency $k$ is 2. Note that this case had already converged at 2 K iterations.

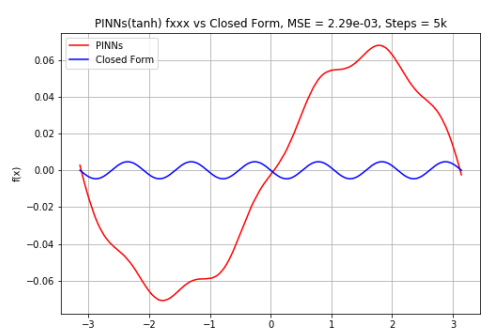

Fig. 19. PINN solution at 5 K iterations of $f_{xxx}(x)$ (eq. (23)) (red) against exact solution (blue), when frequency $k$ is 6. This solution finally converged at 81 k iterations.

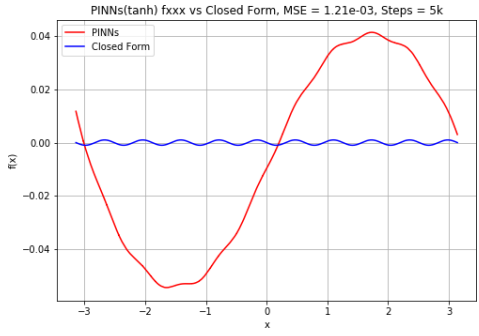 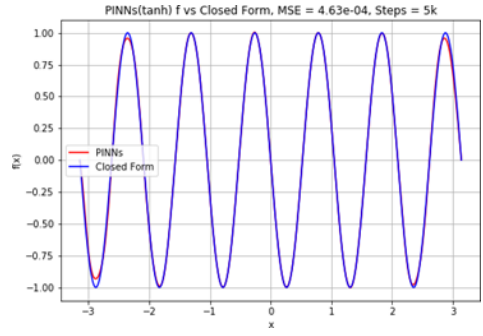

Fig. 20. PINN solution at 5 K iterations of $f_{xxx}(x)$ (eq. (23)) (red) against exact solution (blue), when frequency $k$ is 10. This case did not converge at all.

Fig. 21. PINN solution at 5 K iterations of $f(x)$ (eq. (25)) (red) against exact solution (blue), when frequency $k$ is 6. Comparison with fig. 13 is relevant as it illustrates the causes behind the observations of Table 3, second row.

Carrying forward the discussion on inter-frequency convergences across derivative orders, no clear behavioral pattern is observable till this point. Referring back to eqns. (21-23), it can be seen that while the baseline function (22) comes with a damping coefficient of $1/k^2$ on the RHS, the other two higher-derivative equations have a coefficient of just 1 (i.e., neither damping nor amplification). The next set of investigations checked out if this could have any effect.

The equation solved for is

$$f(x) = -\sin kx, \text{ for } k = 2, 6 \text{ and } 10, \ x \in [-\pi, \pi] \qquad (25)$$

where, compared with (22), the damping coefficient has been removed. The solutions at different frequencies with (25) are compared against solutions obtained with (22). The results are presented in Table 3 below:

Table 3. Convergences of baseline function $f(x)$ of single sinusoids at different frequencies, activation function tanh(.), for PINNs with different coefficients

| Frequency, k | Number of iterations at which convergence is achieved for the function, in thousands (K) | |
|---|---|---|
| | $f(x)$ (eq. 22) (PINN) | $f(x)$, (eq. 25), i.e. normalized coefficients (PINN) |
| 2 | 6 | **2** |
| 6 | 246 | **13** |
| 10 | No convergence | **18** |

The results for normalized coefficients are highlighted in bold and are possibly the most significant finding of the investigations reported in this work. Convergences are significantly improved – across all frequencies – and in fact now we see a clear pattern that is presented in Table 4:

Table 4. Convergences of functions of single sinusoids at different frequencies and derivative orders, activation function tanh(.), when all functions are normalized

| Frequency, k | Number of iterations at which convergence is achieved for the function, in thousands (K) | | |
|---|---|---|---|
| | $f(x)$ (eq. 25) | $f_{xx}(x)$, (eq. 21) | $f_{xxx}(x)$, (eq. 23) |
| 2 | 2 | 2 | 11 |
| 6 | 13 | 18 | 81 |
| 10 | 18 | 50 | No convergence |

The results for normalized Three trends related to PINNs are clearly observable in the above Table 4:
1. First, the more established result, namely that low frequencies are more easily resolved
2. Second, *higher the derivative, the more effort has to be made for attaining converged solution*

3. The variance (spread) in convergence rates across frequencies *increases with increasing orders of the differential equations.*

The last point can be clearly seen in Table 4; the ratio of the number of iterations needed for convergence between the highest and lowest frequencies is 9 for the baseline function, 25 for the 2nd derivative, and indefinitely large for the 3rd derivative.

The above finding overturns the interpretations that were made in Sec. IV.A, namely, that higher derivatives seem to converge faster in PINNs. Even at the stage of coming to this conclusion, there was a discrepancy at the third order equation (20) which did not converge. That equation has high magnitude coefficients, $(2k)^2$, which became very high at the higher frequencies, thus inhibiting convergence. Also, the baseline function (17) coefficients were small. The coefficients were closer to 1 for the first and second derivatives, which exhibited good convergence. So those observations were *related more to the magnitudes of the coefficients associated with different frequency terms at different orders of the differential equations*, than the equation orders themselves

It is pertinent at this point to show one more figure aptly demonstrating the last significant result. Fig. 21 shows solution obtained at 5 K iterations, for frequency k = 6 on eq. (25), that is the baseline function with normalized coefficients.

Finally, we discuss observations from one more experiment, though these are on expected lines. Eq. (22), i.e., the baseline equation with damped coefficients, has been solved using PINNs as reported above. But the derivative-free equation can be naturally solved on fully-connected conventional ANNs in supervised learning mode, MSE cost function, with training data extracted from the closed form. The question is, how does convergence speed compare against the PINN solution, at different frequencies.

The results shown in Table 5 are as expected, i.e., the conventional ANNs are more than one order of magnitude faster. Which implies that, if one needs to create a network for a known equation that is derivative free and has sufficient data to solve in supervised learning mode, there is absolutely no need to use PINNs.

Table 5. Convergences of baseline function *f(x)* of single sinusoids at different frequencies, activation function tanh(.), for PINNs against supervised learning on conventional ANNs with MSE cost function

| Frequency, k | Number of iterations at which convergence is achieved for the function | |
|---|---|---|
| | *f(x)* (eq. 22), for PINN | *f(x)*, (eq. 22), for conventional ANN |
| 2 | 6000 | 440 |
| 6 | 246000 | 2000 |
| 10 | No convergence | 3600 |

The architecture of all ANNs (PINNs and Conventional) considered here used a common 1 x 100 x 100 x 1 pattern, i.e., 2 hidden layers with 100 nodes each. In all cases, only one input node (*x*) and one output (*y* or *f(x)*). Activation function was tanh(.) in all cases except the experiments performed on swish(.). Adam optimizer was used, and learning rate started with 0.005 and gradually reduced with increasing iterations.

Each run was online tracked for convergence of various parameters, both residuals and boundary conditions. Simultaneously the divergence between simulation result and exact solution was also tracked. In almost all cases, even though the convergence parameters reduced to low values, it took more iterations for simulation result to match with exact solution. Only under the latter condition was the solution considered to be converged. The cases reported as "no convergence" were actually found to be diverging with increasing iterations.

Considering the large number of cases, multiple runs for repeatability were made in just a few cases. Only very minor variations were observed.

A clear observation that can be made across all tests performed, is that convergence is best when the coefficient of the forcing term at RHS is close to one. Hence, for drawing conclusions on convergence rates across frequency components and derivative orders, one needs to compare uniformly under this condition. This has been reported exactly in Table 4. There is considerable scope for analysis as to the reasons for the above observations, as well as the conclusion, i.e., higher derivatives need more effort to attain convergence. The work reported here does not get into this analysis.

## V. Conclusions

With the objective of investigating if the phenomena of spectral bias exists in PINNs and its possible variation across orders of differential equations, a series of numerical experiments were conducted on simple sinusoidal functions of different frequencies, compositions and derivative orders.

The conclusions from these experiments are the following:

- The most expected one, that the low frequencies are most easily resolved
- The higher the derivative, the more effort has to be made for attaining converged solution
- As a logical corollaey of the above two, that the variance (spread) in convergence rates across frequencies increases with increasing orders of the differential equations.

## Ackowledgments

The authors acknowledge and express their gratitude for the support received from Dr. Yang Juntao and Dr. Simon See of Nvidia, Singapore, in negotiating some of the aspects of the Modulus framework.